\documentclass[lettersize,journal]{IEEEtran}
\usepackage{amsmath,amsfonts}
\usepackage{algorithmic}
\usepackage{algorithm}
\usepackage{array}
\usepackage{textcomp}
\usepackage{stfloats}
\usepackage{url}
\usepackage{verbatim}
\usepackage{graphicx}
\usepackage{cite}
\usepackage{subfigure}
\usepackage{multirow}
\usepackage{bm}
\usepackage{nicefrac}
\usepackage{makecell}
\usepackage{xcolor}
\usepackage{extarrows}
\usepackage[colorlinks=true,linkcolor=blue,citecolor=blue,urlcolor=purple,]{hyperref}
\begin{document}

\title{Continuous Patch Stitching for Block-wise Image Compression}

\author{Zifu~Zhang, \IEEEmembership{Student Member, IEEE,} Shengxi~Li$^{*}$, \IEEEmembership{Member, IEEE,} Henan~Liu, \IEEEmembership{Student Member, IEEE,}\\ Mai~Xu, \IEEEmembership{Senior Member, IEEE,} Ce~Zhu, \IEEEmembership{Fellow, IEEE}\\\vspace{-2em}
\thanks{Zifu Zhang, Shengxi Li (Corresponding author), Henan Liu and Mai Xu are with the School of Electronic and Information Engineering, Beihang University, Beijing 100191, China (Email: ZifuZhang@buaa.edu.cn; LiShengxi@buaa.edu.cn; Lhn21373089@buaa.edu.cn; MaiXu@buaa.edu.cn). Ce Zhu is with the School of Electronic and Information Engineering, University of Electronic Science and Technology of China, Chengdu 611731, China (Email: eczhu@uestc.edu.cn)}
}
\markboth{Journal of \LaTeX\ Class Files,~Vol.~14, No.~8, February~2025}%
{Shell \MakeLowercase{\textit{et al.}}: A Sample Article Using IEEEtran.cls for IEEE Journals}
\maketitle

\begin{abstract}
Most recently, learned image compression methods have outpaced traditional hand-crafted standard codecs. However, their inference typically requires to input the whole image at the cost of heavy computing resources, especially for high-resolution image compression; otherwise, the block artefact can exist when compressed by blocks within existing learned image compression methods. To address this issue, we propose a novel continuous patch stitching (CPS) framework for block-wise image compression that is able to achieve seamlessly patch stitching and mathematically eliminate block artefact, thus capable of significantly reducing the required computing resources when compressing images. More specifically, the proposed CPS framework is achieved by padding-free operations throughout, with a newly established parallel overlapping stitching strategy to provide a general upper bound for ensuring the continuity.  Upon this, we further propose functional residual blocks with even-sized kernels to achieve down-sampling and up-sampling, together with bottleneck residual blocks retaining feature size to increase network depth. Experimental results demonstrate that our CPS framework achieves the state-of-the-art performance against existing baselines, whilst requiring less than half of computing resources of existing models. Our code shall be released upon acceptance.
\end{abstract}

\vspace{-1em}
\section{Introduction}
\label{sec:intro}
\IEEEPARstart{W}{ith} the rapid development of multimedia technologies, the exponential growth of image and video data volume has imposed significant challenges to transmission bandwidth and storage capability. In response, consecutive image compression standards, including JPEG \cite{rabbani2002overview}, HEVC/BPG \cite{sullivan2012overview}, and VVC all-intra \cite{bross2021overview} codecs, have been developed following block-wise hybrid compression framework, so as to reduce the volume of data while maintaining image quality. Furthermore, the advancement of deep learning techniques allows for the improved compression efficiency by various learned image compression methods, based on either recurrent neural networks \cite{toderici2017full} and convolution neural networks \cite{jiang2017end, minnen2018joint}, in which the attention mechanism has been further employed to capture global redundancy for compression \cite{cheng2020learned, he2022elic}. However, compared to standard block-wise codecs, existing learning-based methods are typically trained in a block-wise style, whereas during the inference stage, the compression is performed upon the whole image, which requires excessive computing resources, especially for high-resolution scenarios. 

Existing methods are seeking for learned block-wise image compression for both training and inference, which has been receiving increasing research efforts in the past decade. Minnen \textit{et al.} \cite{minnen2017spatially} introduced a block-wise compression framework leveraging spatial context prediction for intra-prediction, together with residual encoding through a recurrent autoencoder. Lin \textit{et al.} \cite{lin2020spatial} further employed a hierarchical recurrent architecture to address blocks and sub-blocks, such that the redundancy  across blocks can be further reduced; this exhibits superior performance over both full-resolution {and} HEVC codecs. Most recently, Kamisli \cite{kamisli2024end} proposed a single auto-encoder network with autoregressive block-level masked convolutions, leveraging mutual information between adjacent blocks; this is at the cost of deficiency on compression performance and speed. Despite these advancements, block artefacts persist as a significant challenge, particularly at low bitrates when residuals cannot be accurately predicted across blocks. We also noticed several methods to employ additional post-processing enhancement procedures, by either convolutional filtering \cite{yuan2021block,zhao2021learned,xu2023content} and boundary-aware masks \cite{wu2021learned}; however, those post-processing methods also require the overall input of compressed images, adhere to the path route of existing full-resolution compression paradigm.  Therefore, existing block-wise methods may still suffer from block artefacts, and are also limited to unaffordable computational time and resources.

\begin{figure}[t]
\includegraphics[width = 1\linewidth]{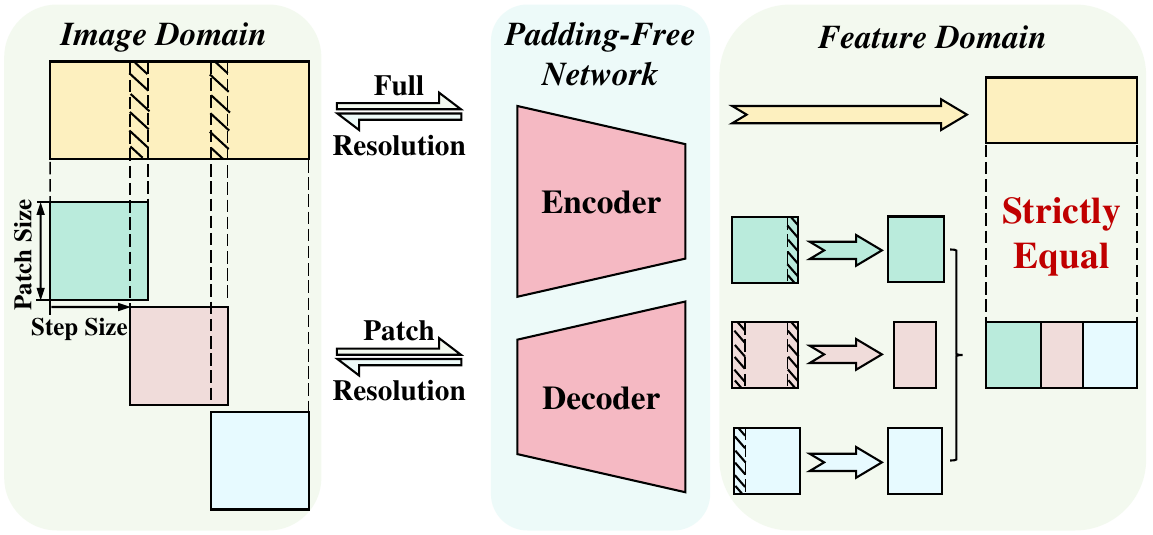}
\vspace{-2em}
\caption{Illustration of the proposed CPS framework, which is able to efficiently compress image patches in parallel, with the mathematical equivalence to compressing a whole image. This also ensures the intrinsic continuity when stitching patches together to an image, even without training our network. }
\vspace{-2.1em}
\label{fig:fig1}
\end{figure}
Instead of compressing blocks in an autoregressive style, we propose in this paper a novel continuous patch stitching (CPS) framework that allows for parallelly connecting image patches at arbitrary locations, even without the training procedure. 
More specifically, by realising the fact that the default convolutional settings with paddings can inevitably lead to the discontinuity when compressing images, we propose to use padding-free setting throughout our CPS framework, and newly establish a parallel overlapping patch stitching (POPS) strategy to provide an upper bound of two consecutive images, such that their continuity after processed by our CPS framework can be mathematically ensured. Under the guidance of the established strategy, we propose functional down-sampling and up-sampling residual blocks, with even-sized convolutional kernels to match input and output patch sizes, the basic requirement for image compression. We also introduce the bottleneck module that retains image patch sizes, with the capability of increasing the receptive field by network depth to improve compression efficiency. Based on the proposed residual blocks, our CPS framework ensures that block artefacts are completely eliminated without the need for additional post-processing operations. Consequently, our CPS framework achieves the state-of-the-art performance for block-wise image compression, with minimal computing resources. For ultra-high-resolution image compression, our CPS framework requires even less than 4GB memory. The proposed CPS framework can also effectively address the out-of-memory issues caused by high-resolution images in other computer vision tasks. We illustrate our CPS framework in Fig. \ref{fig:fig1} and summarize our contributions as follows:
\begin{itemize}
    \item We establish the POPS strategy, which theoretically ensures the intrinsic continuity when compressing images in a block-wise manner.  
    \item We develop the padding-free residual blocks with even-sized kernels, which ensures patch continuity and size match during down-sampling and up-sampling.
    \item We propose the bottleneck residual block under the guidance of the POPS strategy, which retains both continuity and patch sizes for flexibly increasing network depth. 
\end{itemize}
\vspace{-0.5em}
\section{Proposed Method}

\subsection{Parallel Overlapping Patch Stitching (POPS) Strategy}
Padding operation is frequently used in default CNN architectures to prevent the size of feature map from shrinking at each layer, which also allows for the convolution kernel to learn the information at the edge of the image \cite{bhatt2021cnn}. Regarding block-wise compression, the padding operation essentially extends content of image patches, which is inconsistent to the actual content in the extended areas. This leads to the discontinuity at the edge of two consecutive patches. Therefore, to ensure the intrinsic continuity across patches, the prerequisite is the padding-free setting for all convolution layers. 

In our CPS framework, the blocks are encoded independently and thus can be stitched in parallel. Before introducing the detailed architecture, we first introduce the POPS strategy, which ensures that concatenated patch features are mathematically equivalent to those from processing the entire image, as illustrated in Fig. \ref{fig:fig1}. This also strictly ensures the intrinsic continuity across arbitrary two consecutive patches after our CPS framework. 
More specifically, let $r_l$ denote the receptive field, $k_l$ the kernel size, and $s_l$ the stride for the $l$-th layer. The receptive field is calculated by \cite{araujo2019computing}:
\vspace{-0.5em}\begin{equation}
\label{eq:receptive field}
r_l=r_{l-1}+\big((k_l-1)\cdot\prod_{i=1}^{l-1}s_i\big).
\vspace{-0.5em}
\end{equation}
Then, given the symmetry property of the encoder and decoder in compression framework, we assume the input size of image patches as $w_p$, and the overall down-sampling (or up-sampling) factor $\rho_l$ until the $l$-th layer of the encoder (or decoder) network is calculated by $\rho_l=\prod_{i=1}^{l}s_i$. This way, the feature for the $l$-th layer is of the size $\nicefrac{w_p}{\rho_l}$.

\begin{figure}[t]
\includegraphics[width = 1\linewidth]{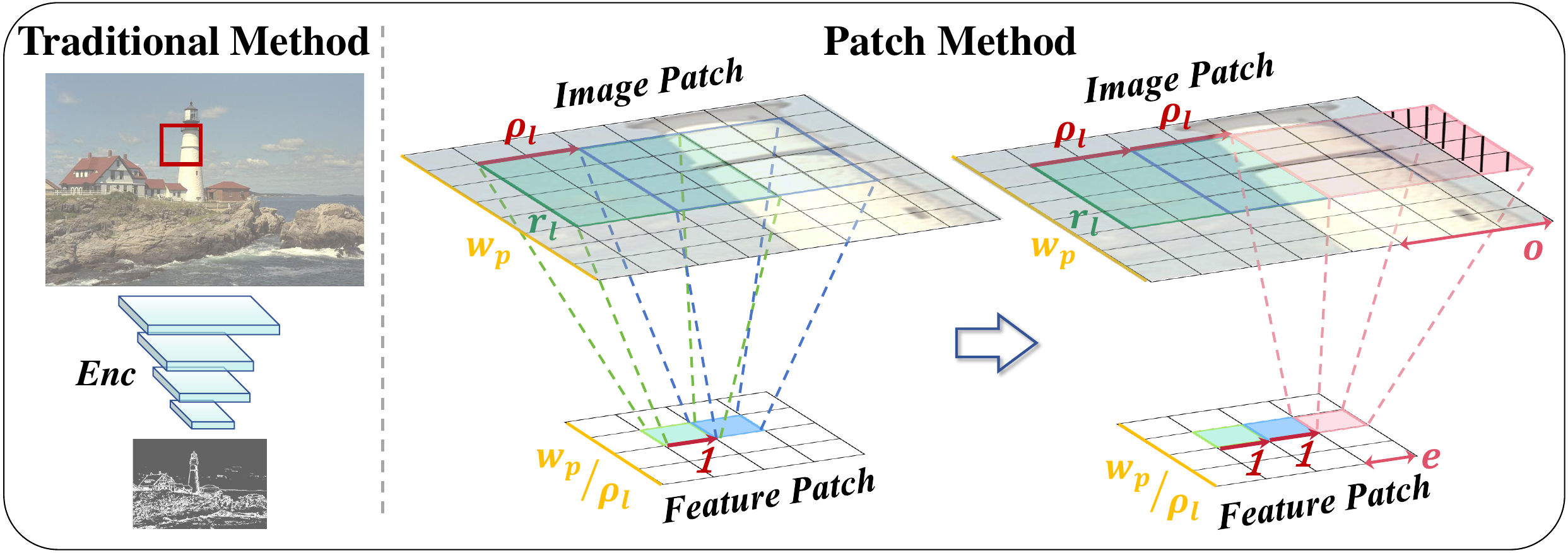}
\vspace{-2em}
\caption{Illustration of our overlapping patch stitching, in which the overlapping size $o$  provides the upper bound of step size to ensure the intrinsic continuity of two consecutive patches, when compressed by our CPS framework. }
\vspace{-1.3em}
\label{fig:fig3}
\end{figure}

Correspondingly, as shown in Fig. \ref{fig:fig3}, from the perspective of the receptive field, shifting one pixel in the feature domain corresponds to $\rho_l$ pixels in the image patch domain. We can thus obtain that $n$ consecutive feature pixels capture $r_l+(n-1)\rho_l$ pixels in the patch domain. In other words, given the image patch of size $w_p$, the valid feature pixels $n_v$ used for reconstruction can be calculated as follows,
\begin{equation}
    r_l+(n-1)\rho_l\leq w_p \rightarrow n_v = \lfloor(w_p - r_l)/\rho_l\rfloor + 1,
\end{equation}
where $\lfloor\cdot\rfloor$ denotes floor operation. Given the feature size $\nicefrac{w_p}{\rho_l}$, we have the unreachable distance $e$ in the feature domain as
\begin{equation}
\label{eq:error in feature domain}
2 e=\frac{w_p}{\rho_l}-(\lfloor\frac{w_p-r_{l}}{\rho_l}\rfloor+1)=\lceil\frac{r_l}{\rho_l}\rceil-1,
\end{equation}
where $\lceil\cdot\rceil$ represents ceil operation. Please note that $2e$ represents that the unreachable distance should be calculated on both sides. From \eqref{eq:error in feature domain}, it is surprising to find that the unreachable region is independent with the input size $w_p$ and solely depends on the receptive field $r_l$ and scale factor $\rho_l$. 

Apparently, the unreachable region is the core that leads to the discontinuity when stitching two patches. We thus propose an overlapping patch stitching strategy in the image domain, such that the unreachable regions can be overlapped between the two consecutive patches. The overlapping size is thus calculated by the scale factor as
\begin{equation}
\label{eq:overlap in image domain}
o=\rho_l\cdot2e=
\left\{
\begin{aligned}
&r_l-\rho_l &\ \mathrm{when}~ r_l \bmod \rho_l = 0 \\
&\rho_l\cdot\lfloor \frac{r_l}{\rho_l}\rfloor &\ \mathrm{when}~ r_l \bmod \rho_l \neq 0\\
\end{aligned}
\right.
\end{equation}
The overlapping size $o$ essentially indicates the maximal step size $(w_p-o)$ between two consecutive patches, which is also shown in Fig. \ref{fig:fig1}. We may also need to point out that the upper bound of step size is quite general and is also applicable to various architectures under the padding-free setting, in both compression and other computer vision tasks.

\vspace{-1em}

\subsection{Padding-free Residual Blocks with Even-sized Kernels}

Since down-sampling and up-sampling are the workhorse to reduce data volume, the convolution and transposed convolution thus act as the crucial role for the encoder and decoder architectures in learned block-wise image compression. Then, although ensuring the intrinsic continuity by our parallel overlapping patch stitching strategy, the padding-free setting with default convolution can lead to another problem, i.e., the mismatch of input and output sizes after the down-sampling and up-sampling procedures. We analyse this by formulating the convolution for each layer  as
\begin{equation}\label{eq:conv io}
    w_{\mathrm{out}} = \lfloor\frac{w_{\mathrm{in}} + 2p - k}{s}\rfloor + 1\xlongequal{p=0}\lfloor\frac{w_{\mathrm{in}} - k}{s}\rfloor + 1,
\end{equation}
where $w_{\mathrm{in}}$ is the input size, $w_{\mathrm{out}}$ is the output size, $k$ denotes the kernel size, $s$ is the stride, and $p=0$ represents the padding-free setting. In contrast, for the transposed convolution, we have the reverse of the input and output sizes, as
\begin{equation}\label{eq:transconv io}
    {w}_{\mathrm{out}} = {s}\cdot({w}_{\mathrm{in}}-1)-2{p}+{k} \xlongequal{{p}=0}{s}\cdot({w}_{\mathrm{in}}-1)+{k}.
\end{equation}
By combining \eqref{eq:conv io} and \eqref{eq:transconv io}, we are able to conclude that $w_{\mathrm{in}}-k$ needs to be divisible by $s$, such that the pair of convolution and transposed convolution can exactly recover the image size, under the padding-free setting.  

\begin{figure}
  \centering
  \subfigure[Down-sampling and Up-sampling]{
    \label{fig:fig2:a} 
    \includegraphics[height=3.65cm]{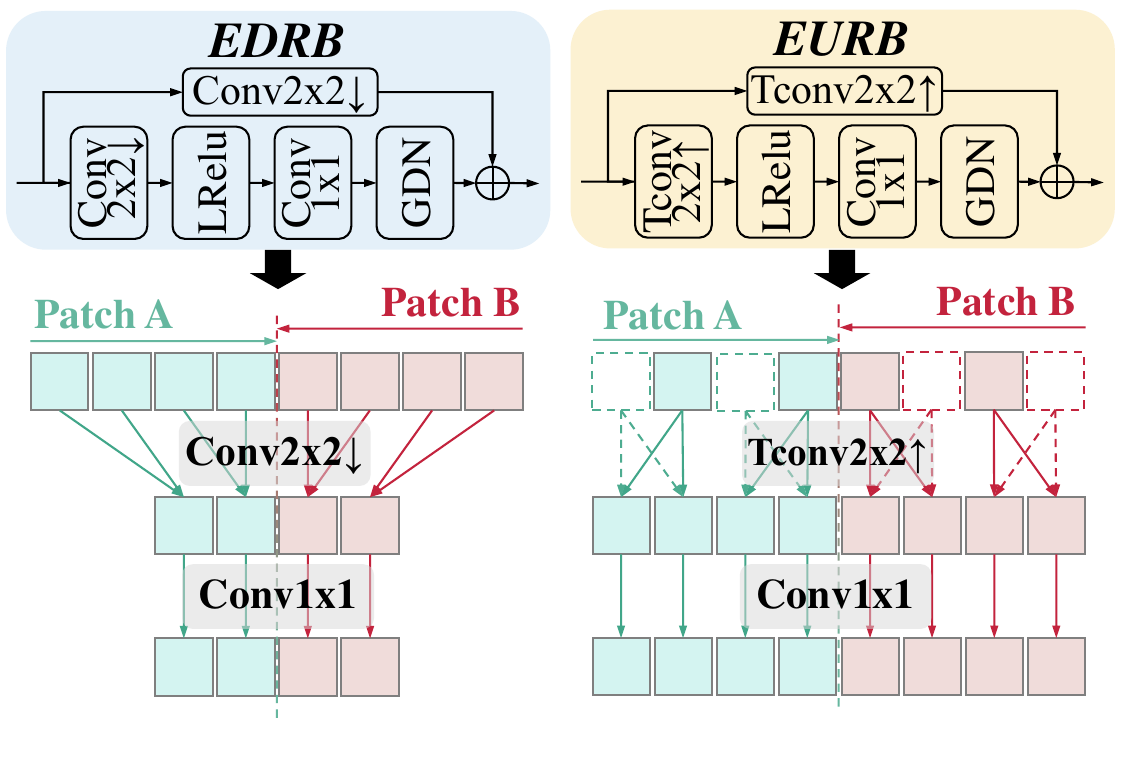}}
  \subfigure[Size Retaining]{
    \label{fig:fig2:b} 
    \includegraphics[height=3.65cm]{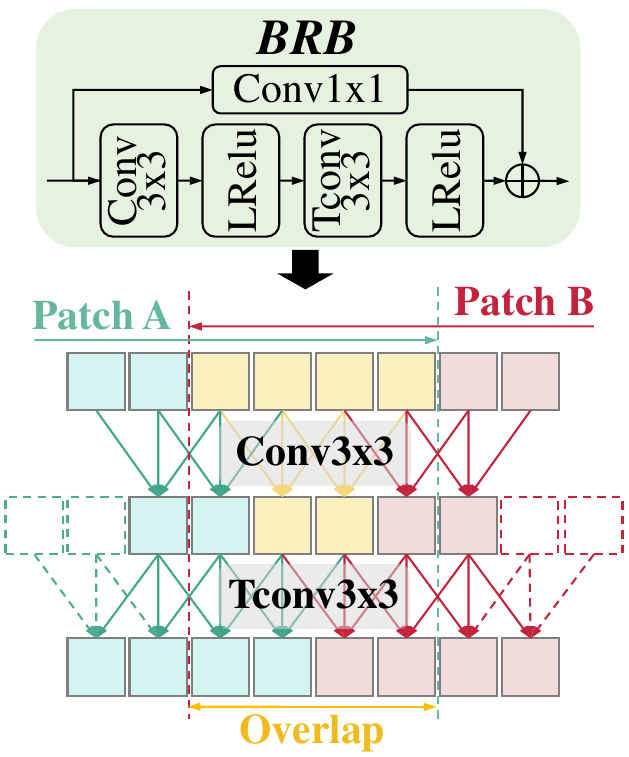}}
  \vspace{-1em}\caption{Basic building blocks of our CPS framework. The arrows indicate up-sampling/down-sampling with a scale of 2. Tconv represents transposed convolution. We also illustrate the impacted areas after important operations, whereby the yellow blocks indicate areas affected by both patches.}
  \label{fig:fig2} 
  \vspace{-1.5em}
\end{figure}

Furthermore, the default convolution (or transposed convolution) sets $s=2$ such that each layer can down-sample (or up-sample) the input size by a ratio of 2. The default odd-sized kernels such as $3\!\times\!3$ and $5\!\times\!5$ settings are thus indivisible by the $s=2$ setting, leading to mismatch between input and output sizes when sequentially processed by the convolution and transposed convolution layers. This mismatch becomes severe when multiple convolution and transposed convolution layers exist, resulting into distinct block artefacts when grouping compressed patches into images. Therefore, given the fact that images to be compressed are typically even-sized, we utilize even-sized kernels (e.g., $k=2$) and establish the even-sized down-sampling residual block (EDRB) and even-sized up-sampling residual block (EURB) as shown in Fig. \ref{fig:fig2:a}. Indeed, two consecutive patches processed by EDRB and EURB retain intrinsic continuity even without overlap according to (\ref{eq:overlap in image domain}), since $r_l=\rho_l$ consistently holds by stacking layers. Moreover, although odd-sized kernels are preferred for their centred and symmetric feature alignment, even-sized kernels are comparably efficient for feature extraction \cite{wu2019convolution}. 


\vspace{-0.5em}

\subsection{Padding-free Bottleneck Residual Block}
Our EDRB and EURB can effectively down-sample and up-sample image patches, with the ability to retain image sizes between input and output patches. However, the learning capacity of networks that rely solely on up-sampling and down-sampling operations is inherently limited. To address this, an efficient way is to increase network depth via stacking modules that are able to retain input and output sizes, so as to increase the receptive field of the network. Though, this is infeasible for convolutions without paddings, except for $k=1$. Although several methods such as shifted convolutions \cite{li2024patch}, have been proposed to address this issue for $1\times1$ convolution, the receptive field still remains highly restricted.

Correspondingly, besides EDRB and EURB, we propose a new bottleneck residual block (BRB) as shown in Fig. \ref{fig:fig2:b}, which consists of convolution and transposed convolution with $s=1$ to ensure the patch size unchanged. Our BRB also employs the $1\!\times\!1$ convolution as the skip connection; this is the standard construction of a residual block. As can be verified via \eqref{eq:conv io} and \eqref{eq:transconv io}, the output feature/patch size remains unchanged for arbitrary $k$. However, our BRB may lead to discontinuity without overlapping. For example, if the input receptive field $r_{0}=1$, we have $k=3$ and $s=1$ in our BRB, outputting the receptive field equal to 5. The overlap size is thus required to 4 to ensure the continuity according to our POPS strategy, which is also shown by the yellow blocks in Fig. \ref{fig:fig2:b}. Consequently, in our CPS framework, EDRB, EURB and BRB essentially constitute all the scenarios required in the network architecture, thus allowing for flexible network depth and efficient representation of spatial information, and eventually improving the compression efficiency by capturing complex patterns and edge details. 


\vspace{-1em}
\subsection{Overall Compression Framework}
\begin{figure}[t]
\centering
\includegraphics[width = 1\linewidth]{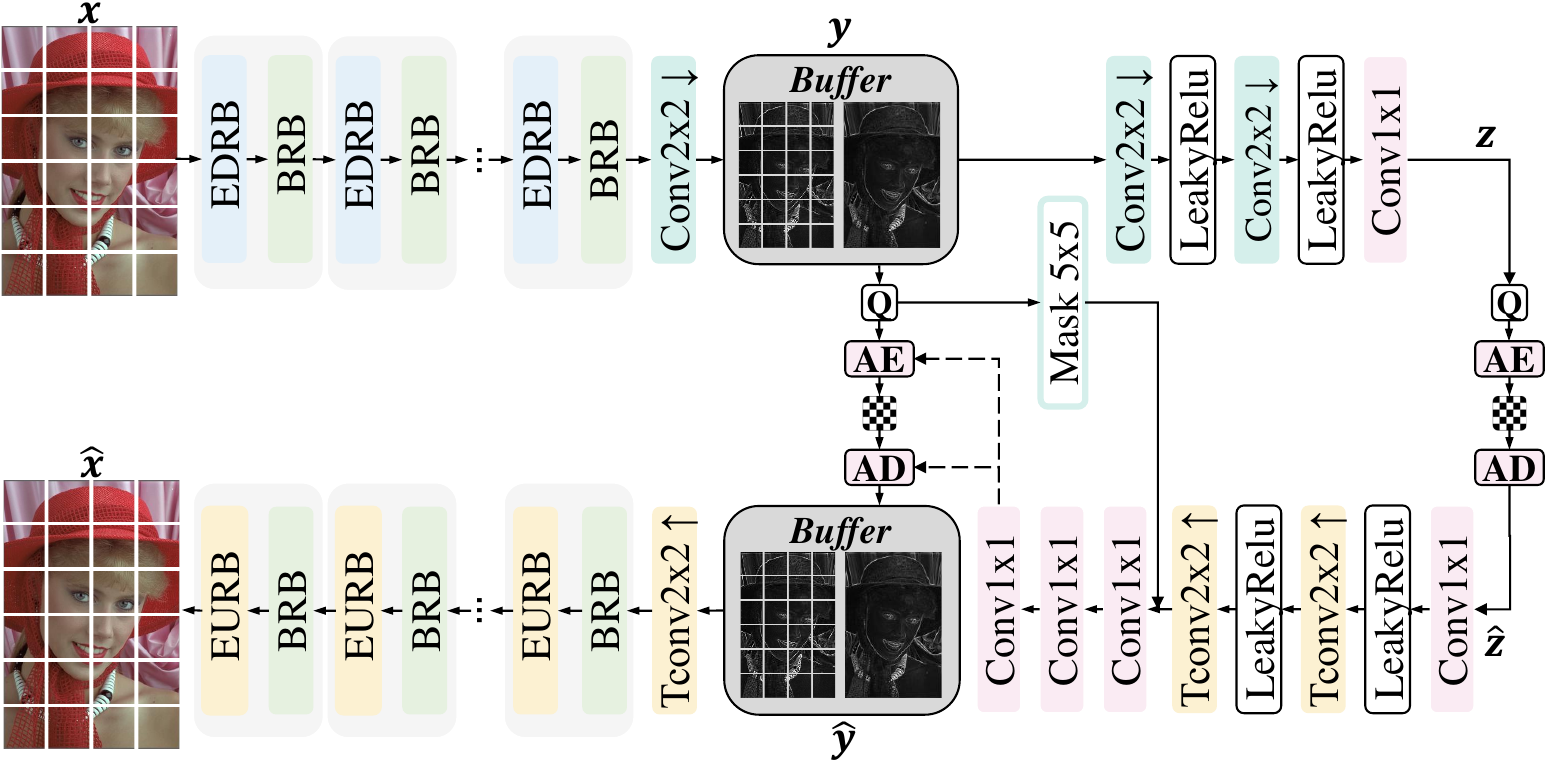}
\vspace{-2em}
\caption{The overall architecture of the proposed network in our CPS framework for block-wise image compression. }
\vspace{-2em}
\label{fig:fig4}
\end{figure}

The architecture of the compression network proposed in our CPS framework is illustrated in Fig. \ref{fig:fig4}. The design of the entropy model is based on the joint autoregressive and hierarchical priors compression network \cite{minnen2018joint}, which is able to effectively exploit the probabilistic structure of latent variables. Each feature element $\bm{y}$ in Fig. \ref{fig:fig4} is modelled as a Gaussian distribution with the mean $\bm{\mu}$ and standard deviation $\bm{\sigma}$, predicted via a parametric hyper autoencoder applied to the hyperprior $\bm{z}$. Since there is no prior knowledge about $\bm{z}$, it is modelled using a fully factorized density model. The rate loss is then calculated from the probability distributions of quantized $\bm{\hat{y}}$ and $\bm{\hat{z}}$ based on Shannon's theory and the mean squared error $\mathcal{D}{(\cdot)}$ is used to measure the quality of the reconstructed patch $\bm{\hat{x}}$, given the original patch $\bm{x}$. Therefore, the loss function is
\begin{equation} \label{eq:loss}
\begin{aligned}
    \mathcal{L}&=\mathcal{R}{(\bm{\hat{y}})}+\mathcal{R}{(\bm{\hat{z}})}+\lambda\mathcal{D}{(\bm{x},\bm{\hat{x}})}\\
    &=\mathbb{E}[-\log_2(p(\bm{\hat{y}}))]+\mathbb{E}[-\log_2(p(\bm{\hat{z}}))]+\lambda\mathcal{D}{(\bm{x},\bm{\hat{x}})}
\end{aligned}
\end{equation}
where $\lambda$ controls the rate-distortion tradeoff. 

\begin{figure}[ht]
\includegraphics[width = 0.97\linewidth]{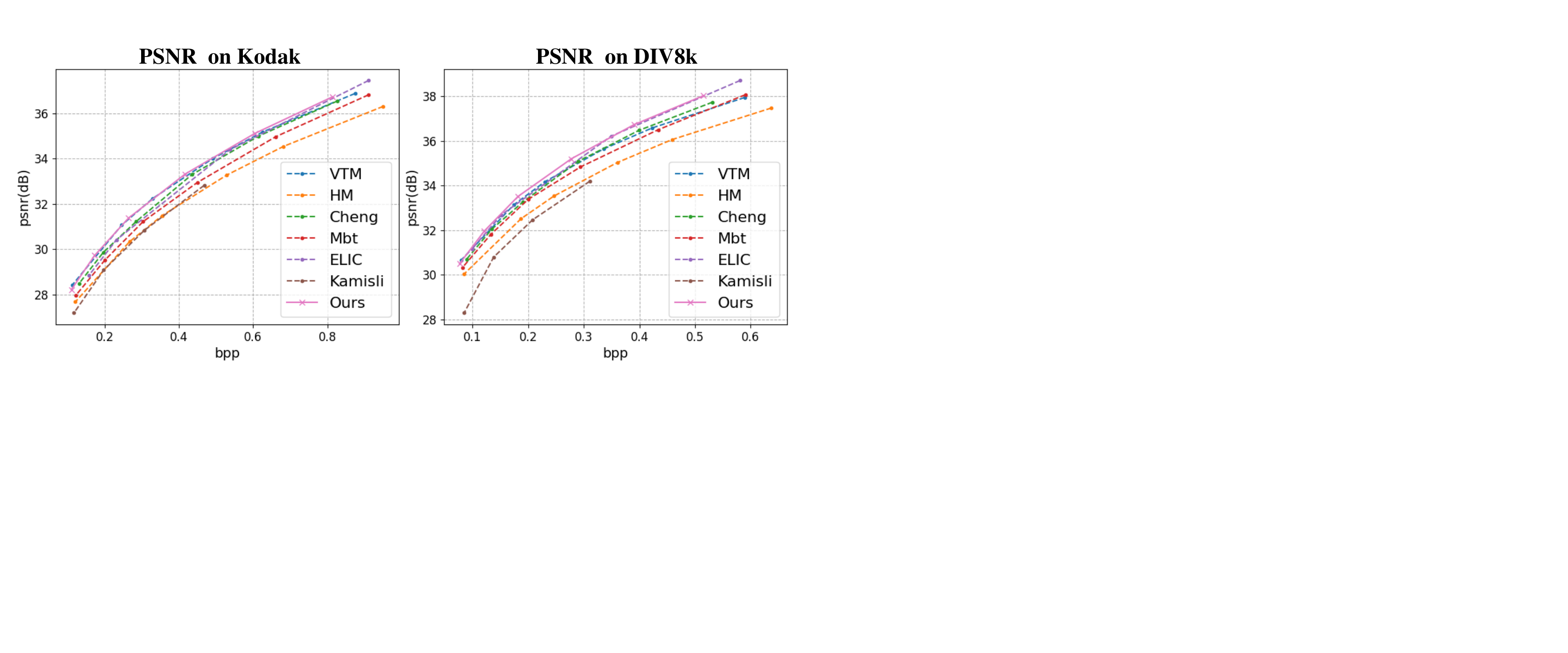}
\vspace{-1em}
\caption{R-D performance covering a wide range of traditional and learned compression methods on Kodak and DIV8k dataset for PSNR.}
\label{fig:fig6}
\vspace{-1.5em}
\end{figure}
\vspace{-1em}
\section{Experiments}
\subsection{Experimental Settings}

\noindent\textbf{Datasets:} For training, we employed the Flicker 2W dataset which is consistent with previous work \cite{liu2020unified}. Each image was randomly cropped into $256\times256$ patches. For testing, we used the Kodak dataset for low-resolution images and the DIV8k dataset \cite{9021973} for high-resolution images. The Kodak dataset consists of 24 lossless images with a resolution of 512 × 768, while the DIV8k dataset includes 24 randomly sampled images with resolutions ranging from 2K to 8K resolutions.

\noindent\textbf{Baseline Methods:} We compared our method with the video coding standards HEVC (HM 16.0\footnote{HM: \href{https://vcgit.hhi.fraunhofer.de/jvet/HM}{https://vcgit.hhi.fraunhofer.de/jvet/HM}}) and VVC (VTM 12.0\footnote{VTM: \href{https://vcgit.hhi.fraunhofer.de/jvet/VVCSoftware_VTM}{https://vcgit.hhi.fraunhofer.de/jvet/VVCSoftware\_VTM}}) using the all-intra configuration with QPs $\{27, 30, 32, 35, 37, 42\}$. For learning-based methods, we evaluated against Mbt \cite{minnen2018joint}, Cheng \cite{cheng2020learned}, and ELIC \cite{he2022elic} using checkpoints from CompressAI\footnote{CompressAI: \href{https://github.com/InterDigitalInc/CompressAI}{https://github.com/InterDigitalInc/CompressAI}}, together with the block-wise method by Kamisli \cite{kamisli2024end}.

\noindent\textbf{Implementation Details:} All experiments were implemented using PyTorch 2.5 and trained with an NVIDIA GeForce RTX 4090 GPU. The training batch size was set to 32 and we implemented a range of $\lambda$ values, including $\{1.8, 3.5, 6.25, 13, 25, 48.3\}\times10^{-3}$ to get the rate-distortion curve. The Adam optimizer was applied with an initial learning rate of $10^{-4}$ for both the main and auxiliary optimizers.  We trained each point for at least $1M$ steps to ensure sufficient convergence. More importantly, we employed {3 pairs of EURB (EDRB) and 6 BRB modules}, and thus according to (\ref{eq:receptive field}), the receptive field of our network is $r_l=72$, with the scale factor $\rho_l=16$, requiring an overlap size of $o=64$ based on (\ref{eq:overlap in image domain}). 
\vspace{-1em}
\subsection{Comparison Results}
\noindent\textbf{Quantitative results.} We evaluated the rate-distortion (RD) performance and the model size against existing learning-based approaches and traditional codecs, as listed in Table \ref{tab:table1}. For fair compression, we applied block-wise compression to all learning-based methods, according to the input image resolution, i.e., 128 patch size for the Kodak dataset and 256 patch size for the DIV8k dataset, except for HM, VTM, and Kamisli methods that are inherently block-wise compression methods. From Table \ref{tab:table1}, our method requires the smallest model size, whilst surpassing all the state-of-the-art baselines particularly for high-resolution images. Fig. \ref{fig:fig6} further plots the RD curves regarding PSNR and MS-SSIM as quality metrics. It can be observed that our CPS framework performs exceptionally well in these two metrics.

\begin{figure}[htbp]
\centering
\includegraphics[width = 0.9\linewidth]{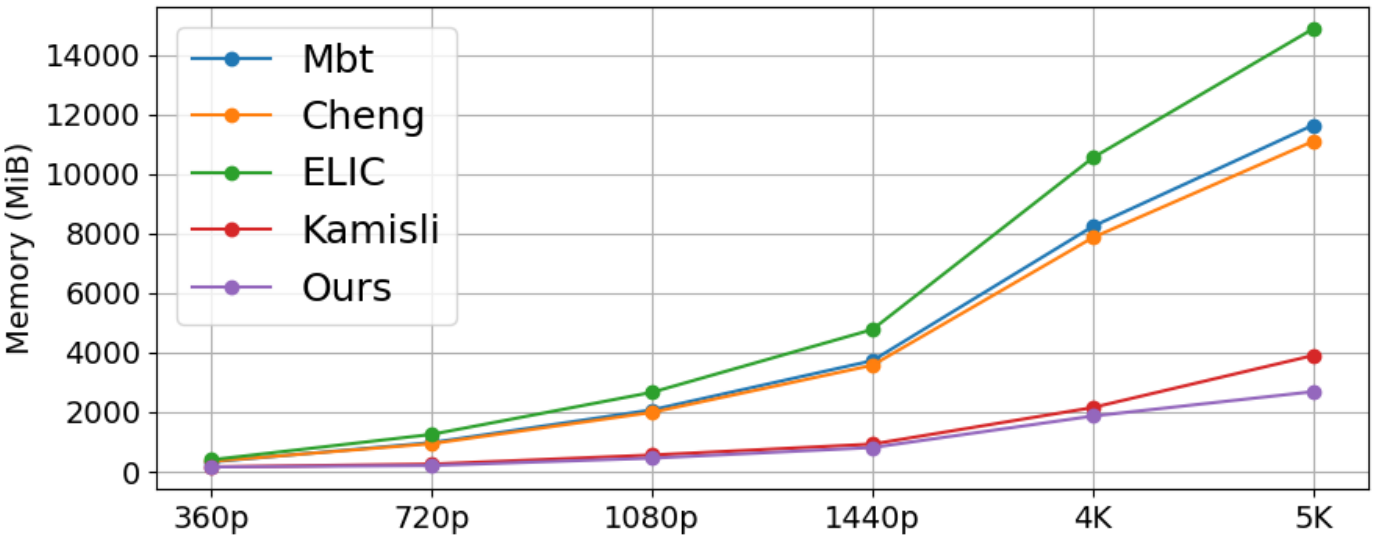}
\vspace{-1em}
\caption{Memory usage (MiB) of learning-based compression methods for images resolutions increasing from 360p (640$\times$360), 720p (1280$\times$720), 1080p (1920$\times$1080), 1440p (2560$\times$1440), 4K (3840$\times$2160), to 5K (5120$\times$2880).}
\label{fig:fig5}
\vspace{-1em}
\end{figure}

\begin{table}[htbp]
\centering
\caption{Comparisons regarding model size and compression efficiency on the Kodak and DIV8k datasets. The BD-rate was calculated based on bpp and PSNR.}
\label{tab:table1}
\begin{tabular}{c|cccc}
\hline
\hline
\textbf{Datasets} & \textbf{Method}  & \textbf{\makecell[c]{Patch\\Size}}  &\textbf{\makecell[c]{Model\\Size (MB)}}& \textbf{BD-rate (\%)}\\ \hline
\multirow{7}{*}{Kodak}        & HM16.0 \cite{sullivan2012overview}           & 64       &   -            & 0.00\\
                              & VTM12.0 \cite{bross2021overview}             & 128      &   -            & -28.38 \\
                              & Mbt \cite{minnen2018joint}               & 128      & 117.65         & -9.68 \\
                            & Cheng \cite{cheng2020learned}              & 128      & 109.35         & -20.09 \\
                             & ELIC \cite{he2022elic}                    & 128      & 137.95         & -20.66 \\
                            &  Kamisli \cite{kamisli2024end}                & 8      & 78.68          & 2.77\\
                            & Ours                                      & 128      & \textbf{43.43} & \textbf{-28.96}\\
\hline
\multirow{7}{*}{DIV8k}         & HM16.0 \cite{sullivan2012overview}          & 64       &   -            & 0.00 \\
                              & VTM12.0 \cite{bross2021overview}             & 128      &   -            & -26.41  \\
                              & Mbt \cite{minnen2018joint}               & 256      & 117.65         & -16.60  \\
                            & Cheng \cite{cheng2020learned}              & 256      & 109.35         & -23.35  \\
                             & ELIC \cite{he2022elic}                    & 256      & 137.95         & -32.54\\
                            &  Kamisli \cite{kamisli2024end}                & 8      & 78.68          & 14.18 \\
                             & Ours                                     & 256      & \textbf{43.43} & \textbf{-34.19} \\
\hline
\hline
\end{tabular}
\vspace{-2.1em}
\end{table}

\begin{figure}[htbp]
\centering
\includegraphics[width = 0.97\linewidth]{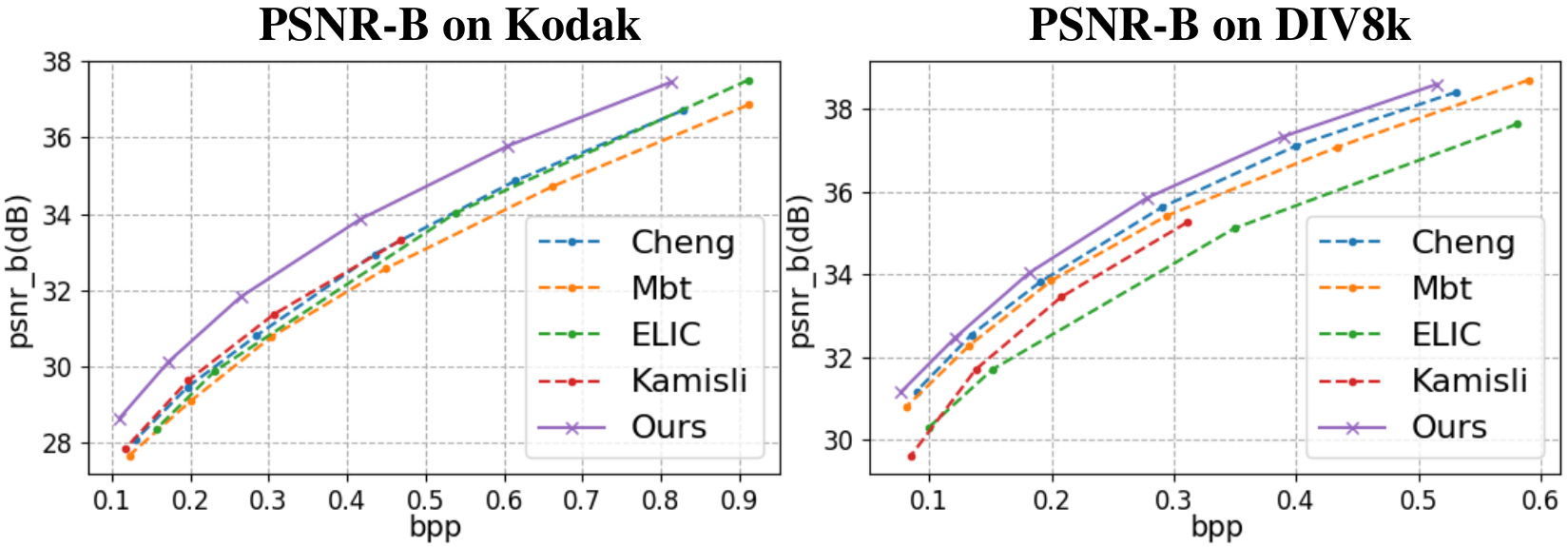}
\vspace{-1em}
\caption{R-D performance on Kodak and DIV8k datasets based on PSNR-B for learning-based compression methods.}
\label{fig:PSNRB}
\vspace{-0.8em}
\end{figure}

Furthermore, we conducted memory usage experiments for images at different resolutions, and reported the results in Fig. \ref{fig:fig5}. With increasing image size, the memory usage of full-resolution learned image compression methods grows exponentially, whereas our CPS framework maintains stable memory usage for seamlessly patch stitching. We further evaluate the block artefacts based on the widely applied PSNR-B metric, and report the result in Fig. \ref{fig:PSNRB}, which verifies the seamless block-wise compression of our CPS framework.

\vspace{-1em}

\begin{figure*}[htbp]
\centering
\includegraphics[width = 0.925\linewidth]{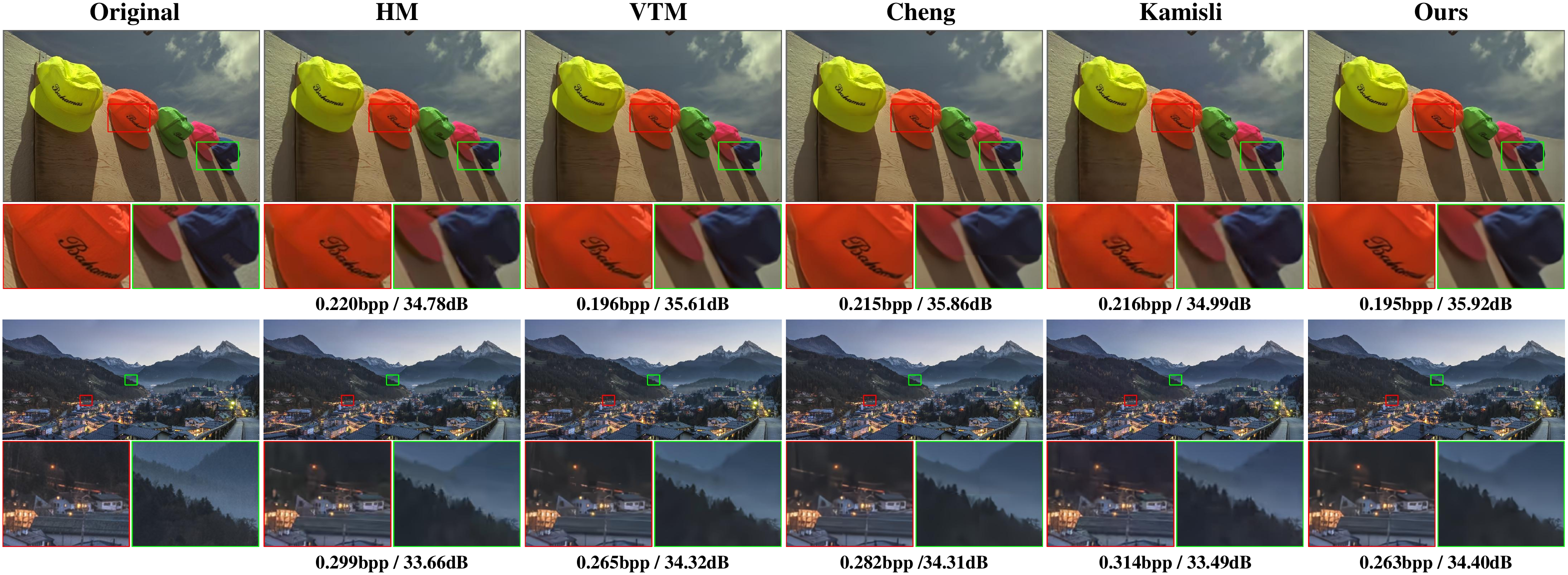}
\vspace{-1em}
\caption{Qualitative comparisons of HM, VTM, Cheng \cite{cheng2020learned}, Kamisli \cite{kamisli2024end}, and our CPS methods, on the Kodak and DIV8k dataset in terms of bpp/PSNR.}
\label{fig:fig7}
\vspace{-1.5em}
\end{figure*}

\noindent\textbf{Qualitative results.} We also provide subjective comparison of our CPS framework with traditional codecs (including HM 16.0 and VTM 12.0) and 2 representative learning-based image compression methods in Fig. \ref{fig:fig7}. As can be seen from this figure, visual artefacts such as blurring and ringing are noticeable in the images compressed by traditional codecs. The learning-based methods, even for the block-wise method, exhibit severe artefacts such as block artefacts. In contrast, our method, leveraging our POPS strategy and specially designed residual blocks, completely eliminates block artefacts and achieves superior visual quality by the lowest bit rates.

\vspace{-1em}
\subsection{Ablation Study}

We further evaluated the core components and settings in our CPS framework, by ablating on the network channel numbers, patch sizes, and kernel sizes of the original $3\times3$ convolution and transposed convolution layers in our BRB. We report the results in Table \ref{tab:table2}, based on the model trained with $\lambda=0.013$. Observe that the 192-channel model achieves saturation, outperforming the 128-channel model, while the 256-channel model does exhibits significant improvement. Moreover, our BRB  with $3\times3$ kernel size observes the significant performance improvement for compression. Our CPS framework is also robust to patch sizes, in which we witness neglectable impact on the performance; this enables the patch size to be flexibly adjusted upon memory available.
\vspace{-2em}
\begin{table}[htbp]
\centering
\caption{Ablation study on core components.}
\label{tab:table2}
\begin{tabular}{ccccc}
\hline
\hline
\textbf{Patch Size} & \textbf{Channels}  & \textbf{BRB Kernel Size}  &\textbf{Bpp}   & \textbf{PSNR}\\ 
\hline
 128          & 128      & $3\times3$     &0.417     &32.92 \\
 128          & 192      & $1\times1$     &0.477      &32.93 \\
 128          & 256      & $3\times3$     &0.416      &33.31 \\
 256          & 192      & $3\times3$     &0.417      &\textbf{33.32} \\
 128          & 192      & $3\times3$     &0.417      &\textbf{33.32} \\
\hline
\hline
\vspace{-3em}
\end{tabular}
\end{table}
\vspace{-0.5em}
\section{Conclusion}

In this paper, we have proposed a novel block-wise image compression method that is able to intrinsically achieve continuous patch stitching, named as CPS framework. The continuity has been guaranteed by our parallel overlapping patch stitching (POPS) strategy that provides a general upper bound to ensure the continuity of two consecutive patches. Under the guidance of the POPS strategy, we have introduced the even-sized down-sampling and up-sampling residual blocks, in particular for matching input and output patch sizes for compression. We also proposed the bottleneck residual block that is able to retain patch size and thus increase network depth for improved compression efficiency. Experimental results have verified the superior performances of our CPS framework, regarding both memory usage and bitrate-distortion performances. Note that our CPS framework was able to handle ultra-high-resolution image compression with less than 4GB memory. Our CPS frameworks has been shown to completely eliminate block artefacts without the need for post-processing,  with the mathematical equivalence to full-resolution image compression; this is highly applicable to other distributed computer vision tasks.

\small{
\bibliographystyle{IEEEtran}

\bibliography{references}}

\end{document}